\documentclass[runningheads,a4paper]{llncs}
\usepackage{amssymb}
\usepackage{amsfonts}
\usepackage{amsmath}
\usepackage{graphicx}
\usepackage{makeidx}  
\usepackage{url}
\newcommand{\keywords}[1]{\par\addvspace\baselineskip\noindent\keywordname\enspace\ignorespaces#1}
\begin{document}
\mainmatter  

\title{A Cascade Neural Network Architecture investigating Surface
  Plasmon Polaritons propagation for thin metals in OpenMP}

\author{\IEEEauthorblockN{Christian
Napoli\IEEEauthorrefmark{1,*},
Giuseppe Pappalardo \IEEEauthorrefmark{1}, and
Emiliano Tramontana\IEEEauthorrefmark{1}}
\IEEEauthorblockA{\IEEEauthorrefmark{1}Dpt. of Mathematics and
Computer Science, University of Catania, Italy}
an unwanted space
\thanks{*Email: napoli@dmi.unict.it.}
\thanks{ 13th International Conference on Artificial Intelligence and Soft Computing (ICAISC) 2014}}

\markboth{A Cascade Neural Network Architecture investigating Surface
  Plasmon Polaritons propagation for thin metals in OpenMP -- PREPRINT}%
{Shell \MakeLowercase{\textit{et al.}}: Bare Demo of
IEEEtran.cls for Journals}

 \begin{titlepage}
 \begin{center}
 {\Large \sc PREPRINT VERSION\\}
  \vspace{5mm}
{\huge A Cascade Neural Network Architecture investigating Surface
  Plasmon Polaritons propagation for thin metals in OpenMP\\}
 \vspace{10mm}
 {\Large F. Bonanno, G. Capizzi, G. Lo Sciuto, C. Napoli, G. Pappalardo, and E. Tramontana\\}
 \vspace{5mm}
{\Large \sc TO BE PUBLISHED ON: \bf 13th International Conference on Artificial Intelligence and Soft Computing (ICAISC) 2014\\}
 \end{center}
 \vspace{5mm}
 {\Large \sc BIBITEX: \\}
 
@incollection\{\\
year=\{2014\},\\
isbn=\{978-3-319-07172-5\},\\
booktitle=\{Artificial Intelligence and Soft Computing\},\\
volume=\{8468\},\\
series=\{Lecture Notes in Artificial Intelligence\},\\
editor=\{Rutkowski, Leszek and Korytkowski, Marcin and Scherer, Rafa and Tadeusiewicz, Ryszard and Zadeh, LotfiA. and Zurada, Jacek M.\},\\
title=\{A Cascade Neural Network Architecture investigating Surface Plasmon Polaritons propagation for thin metals in OpenMP\}, \\
publisher=\{Springer Berlin Heidelberg\},\\
author=\{Bonanno, Francesco and Capizzi, Giacomo and Lo Sciuto, Grazia and Napoli, Christian and Pappalardo, Giuseppe and Tramontana Emiliano\},\\
pages=\{22-33\}\\
\}
 \vspace{5mm}
 \begin{center}
Published version copyright \copyright~2014 SPRINGER \\
\vspace{5mm}
UPLOADED UNDER SELF-ARCHIVING POLICIES\\
NO COPYRIGHT INFRINGEMENT INTENDED \\
 \end{center}
\end{titlepage}

\title{A Cascade Neural Network Architecture investigating Surface
  Plasmon Polaritons propagation for thin metals in OpenMP}
%
\titlerunning{A Cascade NN finding SPPs propagation}
%
%
\author{Francesco Bonanno\inst{1}\and Giacomo Capizzi\inst{1}\and
  Grazia Lo Sciuto\inst{2} \and Christian Napoli\inst{3}\and Giuseppe Pappalardo\inst{3}\and Emiliano Tramontana\inst{3}}
\authorrunning{Bonanno, Capizzi, Lo Sciuto, Napoli, Pappalardo,
  Tramontana}
\institute{Dpt. of Electric, Electronic and Informatics Eng., University of Catania, ITALY
\email{gcapizzi@diees.unict.it}
\and Department of Industrial Engineering, University of Catania, ITALY
\email{glosciuto@dii.unict.it}
\and Department of Mathematics and Informatics, University of Catania, ITALY
\email{\{napoli, pappalardo, tramontana\}@dmi.unict.it}
}


\maketitle

\begin{abstract}
  Surface plasmon polaritons (SPPs) confined along metal-dielectric
  interface have attracted a relevant interest in the area of
  ultracompact photonic circuits, photovoltaic devices and other
  applications due to their strong field confinement and enhancement.
  This paper investigates a novel cascade neural network (NN)
  architecture to
  find the dependance of metal thickness on the SPP propagation.
  Additionally, a novel training procedure for the proposed cascade NN
  has been developed using an OpenMP-based framework, thus greatly
  reducing training time.
  The performed experiments confirm the effectiveness of the proposed NN
  architecture for the problem at hand.
\end{abstract}

\section{Introduction}

Surface Plasmon Polaritons (SPPs) are quantized charge density
oscillations occurring at the interface between a metal and a
dielectric when a photon couples to the free electron gas of the
metal. The emerging field of surface plasmonics has applied SPP
coupling to a number of new and interesting applications
\cite{Franken},\cite{Atwater},\cite{Fahr}, such as Surface Enhanced
Raman Spectroscopy (SERS), photovoltaic devices optimisation, optical
filters, photonic band gap structures, biological and chemical
sensing, and SPP enhanced photodetectors.

Some papers appeared in literature simulate and analyse the excitation
and propagation of SPPs on sinusoidal metallic gratings in conical
mounting.  Researchers working in the emerging field of plasmonics
have shown the significant contribution of SPPs for applications in
sensing and optical communication.

One promising solution is to fabricate optical systems at
metal-dielectric interfaces, where electromagnetic modes called SPPs
offer unique opportunities to confine and control light at length
scales below $100 \,nm$ \cite{Walters},\cite{Waele}.

The studies and experiences conducted on SPPs are well assessed and
show that the propagation phenomena are well established by the
involved materials in the plasmon structure at large thickness,
conversely when it becomes smaller than the wavelength of the
exciting wave, investigations are required due to the actual
poor understanding~\cite{Shah}.

This paper proposes a novel neural netwok (NN) topology for the study
of the problems of a SPP propagating at a metal flat interface
separating dielectric medium.  Currently, we are using NNs to study
the inner relation between SPPs exciting wavelength, metal thickness
and SPP wavelength and propagation length.  The focus of this paper is
on the determination of the dependance of the SPP propagation of the
metal thickness by recurring to suitable NN schematics.
Due to the high sensitivity of the neural model to data oscillations a
novel training procedure has been devised in order to avoid
polarisations and miscorrections of some NN weights.  Moreover, since
such a training procedure could be expensive in terms of computational
power and wall-clock time, a parallel version using an OpenMP
environment, with shared memory, has been developed and optimised to
obtain maximum advantage from the available parallel hardware.
A big amount of data has been put into proper use for the investigated
NN topology. Such data have been made available by the resolution of
3D Maxwell equations with relative boundary conditions performed by
COMSOL Multiphysics, which is an efficient and powerful software
package to simulate the characteristics of SPPs.

\section{Basics of Surface Plasmon Polaritons}
\label{polaritoni}

The field of plasmonics is witnessing a growing interest with an
emerging rapid development due to the studies and researches about the
behaviour of light at the nanometer scale. Light absorption by solar
cells patterned with metallic nanogratings has been recently
investigated, however we consider light-excited SPPs at the metal
surface.  The outcomes of our investigation can be used to improve
efficient capturing of light in solar energy conversion
cells~\cite{Franken}.  Therefore, our main research interests are
toward the properties of SPPs.

SPPs are electromagnetic waves propagating along metal-dielectric
interfaces and exist over a wide range of frequencies, evanescently
decaying in the perpendicular direction.  Such electromagnetic surface
waves arise via the coupling of the electromagnetic fields to
oscillations of the conductor electron's plasma \cite{Maier}.  SPP is
the fundamental excitation mode at a metal-dielectric interface that
is coupled to an electromagnetic wave as described by \cite{Maier}.
The most simple geometry sustaining SPPs is that of a single, flat
interface (see Fig.~\ref{s1}) between a dielectric, non-absorbing half
space ($z > 0$) with positive real dielectric constant $\varepsilon_2$
and an adjacent conducting half space ($z <0$) described via a
dielectric function $\varepsilon_1(\omega)$.  The requirement of
metallic character implies that $Re [\varepsilon_1] < 0$.  As shown in
\cite{Maier}, for metals this condition is fulfilled at frequencies
below the bulk plasmon frequency $\omega_p$. We look for propagating
wave solutions confined to the interface, i.e. with evanescent decay
in the perpendicular z-direction \cite{Maier}. 

The electromagnetic field of a SPP at the dielectric-metal interface
is obtained by solving Maxwell's equations in each medium with the
associated boundary conditions.  The adopted structure is a
metal-dielectric interface composed by Molybdenum and air as shown in
Fig.~\ref{s1}. This structure is the most simple in order to reduce
computational effort, as the main purpose of the paper is to
investigate the important relation between dispersion and thickness of
the metal by means of a proper novel NN architecture.  It should be
noted that this relation is not affected by the complexity of the
structure.

The basic mathematical equations describing the electromagnetic
phenomena concerning SPP propagation are listed below:

{\setlength\arraycolsep{0.2em}
\begin{equation}
\begin{array}{lcl}
\mathbf{H}_{d} & =&\left(0, H_{yd}, 0 \right) e^{i(k_{xd}\,x+k_{zd}\,z-\omega t)} \\
\mathbf{E}_{d}  &=&\left(E_{xd}, 0, E_{zd} \right) e^{i(k_{xd}\,x+k_{zd}\,z-\omega t)}\\
\mathbf{H}_{m} &=&\left(0, H_{ym}, 0 \right) e^{i(k_{xm}\,x-k_{zm}\,z-\omega t)} \\
\mathbf{E}_{m} &=&\left(E_{xm}, 0, E_{zm} \right) e^{i(k_{xm}\,x-k_{zm}\,z-\omega t)}
\end{array}
\label{eq:1}
\end{equation}
}
with boundary condition at $z=0$

{\setlength\arraycolsep{0.2em}
\begin{equation}
\begin{array}{rcl}
 E_{xm}  &=&  E_{xd}\\
H_{ym} &=& H_{yd}\\
\varepsilon_{m}\, E_{zm}  &=& \varepsilon_d\, E_{zd}
\end{array}
\label{eq:2}
\end{equation}
}
as a consequence of the previous equation we have

\begin{equation}
k_{xm}  = k_{xd}
\label{eq:3}
\end{equation}

We consider a system consisting of a dielectric material,
characterised by an isotropic, real, positive dielectric constant
$\varepsilon_d$, and a metal characterised by an isotropic, frequency
dependent, complex dielectric function $ \varepsilon_m =
\varepsilon_r+i\varepsilon_i$.  In order to introduce the main
parameters characterising SPPs assuming the interface is normal to
z-axis and the SPPs propagate along the x direction (i.e., $k_y = 0$),
the SPP wavevector $k_x$ or $\beta$ is related to the optical
frequency $\omega $ through the dispersion relation.

\begin{equation}
k_x= k_0\, \sqrt{\frac{\varepsilon_d\,\varepsilon_m}{\varepsilon_d+\varepsilon_m}}
\label{eq:4}
\end{equation}

\begin{equation}
\beta= \frac{\omega}{c}\, \sqrt{\frac{\varepsilon_d\,\varepsilon_m}{\varepsilon_d+\varepsilon_m}}
\label{eq:5}
\end{equation}

We take $\omega $ to be real and allow $k_x$  to be complex, since our main interest is in stationary monochromatic SPP fields in a finite area, where 

\begin{equation}
k_0=\frac{\omega}{c}
\label{eq:6}
\end{equation}
is the wavevector in free space, and $\lambda_0=\frac{c}{\omega}$ is
the wavelength in vacuum. For metals, the permittivity is complex,
which leads to $k_x$ being complex. The imaginary part of $k_x$
defines the SPP’s damping and as it propagates along the surface.  The
real part of $k_x$  is connected to the plasmon’s wavelength,
$\lambda_{SPP} $: 

\begin{equation}
\lambda_{SPP} =\frac{2\pi}{Re[\beta]}
\label{eq:7}
\end{equation}
$L_{SPP}$ is the SPP propagation length, physically the energy
dissipated through the metal heating and it is the propagation
distance.
$L_{SPP}$ is defined as follows: 

\begin{equation}
L_{SPP} =\frac{1}{Im [\beta]}
\label{eq:8}
\end{equation}

Finally,  the following reports the expression of the electric field of plasmon wave:

\begin{equation}
\mathbf{E}_{SPP} =\mathbf{E}_{0}^{\pm}\, e^{i(k_{x}x\pm k_{z}z-\omega t)} 
\label{eq:9}
\end{equation}
where
\begin{equation*}
\begin{array}{rcl}
k_x &=&k_{x}^{'}+ ik_{x}^{''} \\
k_{x}^{'} &=&\frac{2\pi}{\lambda_{SPP}}
\end{array}
\end{equation*}

\section{Input Data for the Proposed Cascade NN
  Architecture} 

\begin{figure}[!t]
\centering
\begin{minipage}[c]{.49\textwidth}
\centering
\hfill\includegraphics[width=\textwidth]{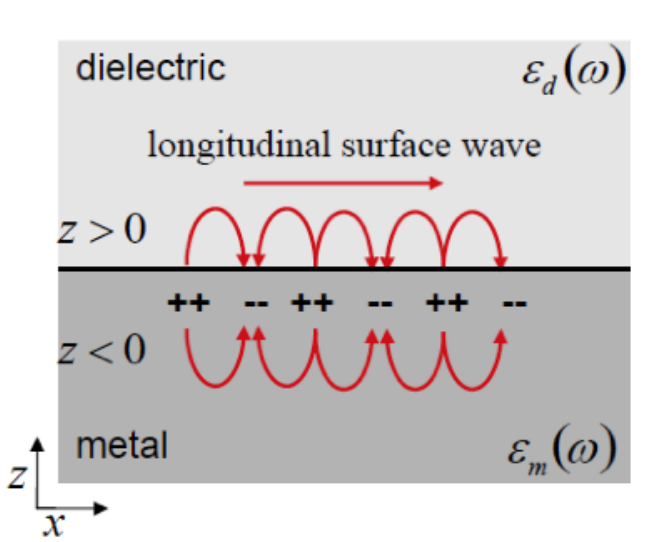}
\caption{Geometry for SPP propagation at a single interface between
  metal and  dielectric.}
\label{s1}
\end{minipage}
~~
\begin{minipage}[c]{.46\textwidth}
\centering
\hfill\includegraphics[width=\textwidth]{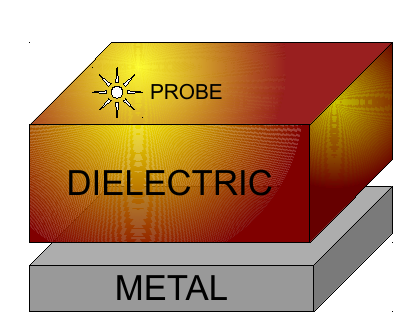}
\caption{Implemented geometry in COMSOL.}
\label{s2}
\end{minipage}
\end{figure}

By solving the full wave 3D Maxwell equations  in the simple geometry
shown in Fig.~\ref{s2},  which separates two media as metal and
dielectric, using the finite element method-based software package
COMSOL Multiphysics,  we have obtained the $L_{SPP}$ and
$\lambda_{SPP} $ data values for different thickness  values.  The
perfectly matched  layer boundary condition was chosen for the
external surface of the plasmon structure. The exciting wave was
monochromatic on the visible  spectra and ranging from $400 \,nm$ to
$700 \,nm$.  

We have performed many numerical simulations while varying the
exciting wavelengths for each investigated thickness, hence obtaining
the corresponding SPP waves.  A SPP propagates at the interface
dielectric-metal decaying into the metal.

The values of $L_{SPP}$ and $\lambda_{SPP} $ were computed for the all
visible range of wavelength at the following different thickness
values $t$ of the metal: $36 \,nm$, $42 \,nm$, $48 \,nm$, $54 \,nm$,
$60 \,nm$, $72 \,nm$, $84 \,nm$, $96 \,nm$ and $128 \,nm$.

\section{The Proposed Neural Network Architecture}

The prediction of $\lambda_{SPP}$ and $L_{SPP}$ from the set of values
$\lambda_0$ and $t$ is related to the problem of the dependence of
$L_{SPP}$ from $\lambda_{SPP}$.  To obtain a correct prediction of
$\lambda_{SPP}$ by a neural network-based approach a value of
$\lambda_{SPP}$ is needed.  Although this can be obtained by a cascade
process, the traditional means have that the NN cascade is
accommodated by separate training sessions for each different
dedicated NN. Unfortunately, such training sessions would result
in very time-consuming computation.
 
In order to overcome the said complication, this paper proposes a
novel parallel paradigm for training that 
manages to run a single comprehensive training for the NN cascade as a
whole, thus avoiding separate training phases. This novel solution has
been used for the problem at hand, described in Section~\ref{polaritoni}.

Essentially, the adopted topology has been derived from a pair of
common two-layer feed-forward neural networks (FFNNs) \cite{Mandic}, used to
separately predict $\lambda_{SPP}$ and $L_{SPP}$, respectively.  The
comprehensive structure is similar to a cascade feed-forward, whereby
the output of the first neuroprocessing stage is connected with the
input of the second stage
and form a new extended input vector for the second stage.
On the other hand, the vector provided as input to the second
neuroprocessing stage depends on the predicted values obtained from
the first stage, hence it propagates a prediction error.  

Moreover, during the training phase, while some outputs can be
validated for the first neuroprocessing stage, the localised deviation
from the correct frequency spectrum could corrupt the training of the
second stage. The behaviour of this novel topology is as a two step
processing of the data signal that is comprehensive also of a so
called \emph{second validation} or \emph{$\omega$-validation},
described in the following, aiming at avoiding such an error
propagation, which would otherwise endanger the correct training of the second
neuroprocessing stage.

A given output from the first stage has to be validated on the
frequencies domain, by a validation module, before it can be
used. This validation module performs the $\omega$-validation by means
of the Fourier computation on a delayed Gaussian window of the output
and training signal.

An intermediate level of data processing requires the implementation
of a module performing the Fourier transform of the data. Its relative
parameters are not \emph{a priori} established, however are on-line
determined by the novel NN topology and then by its training
procedure. 

\begin{figure}[!t]
  \centering
    \begin{minipage}[l]{.55\textwidth}
      \centering
      \hfill\includegraphics[width=\textwidth]{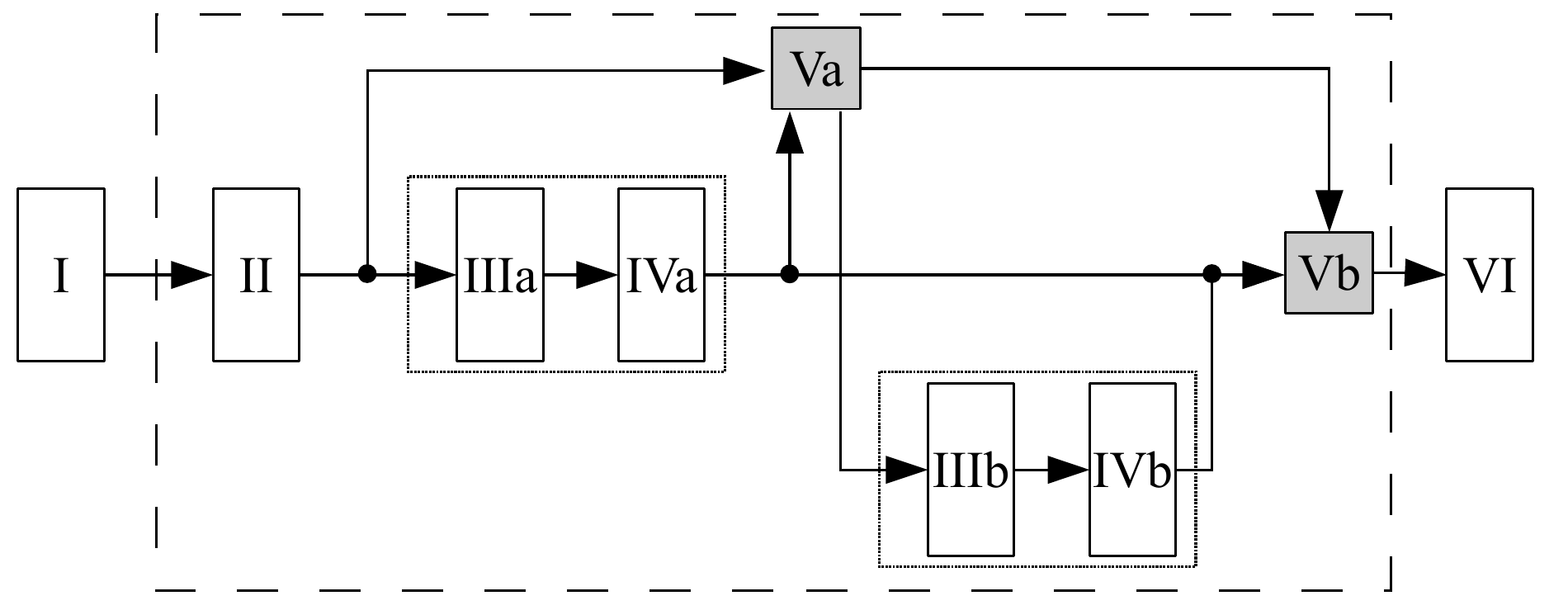}
      \caption{The proposed cascade NN architecture}
      \label{fig:cascade}
    \end{minipage}
    ~~
    \begin{minipage}[r]{.43\textwidth}
      \centering
      \hfill\includegraphics[width=\textwidth]{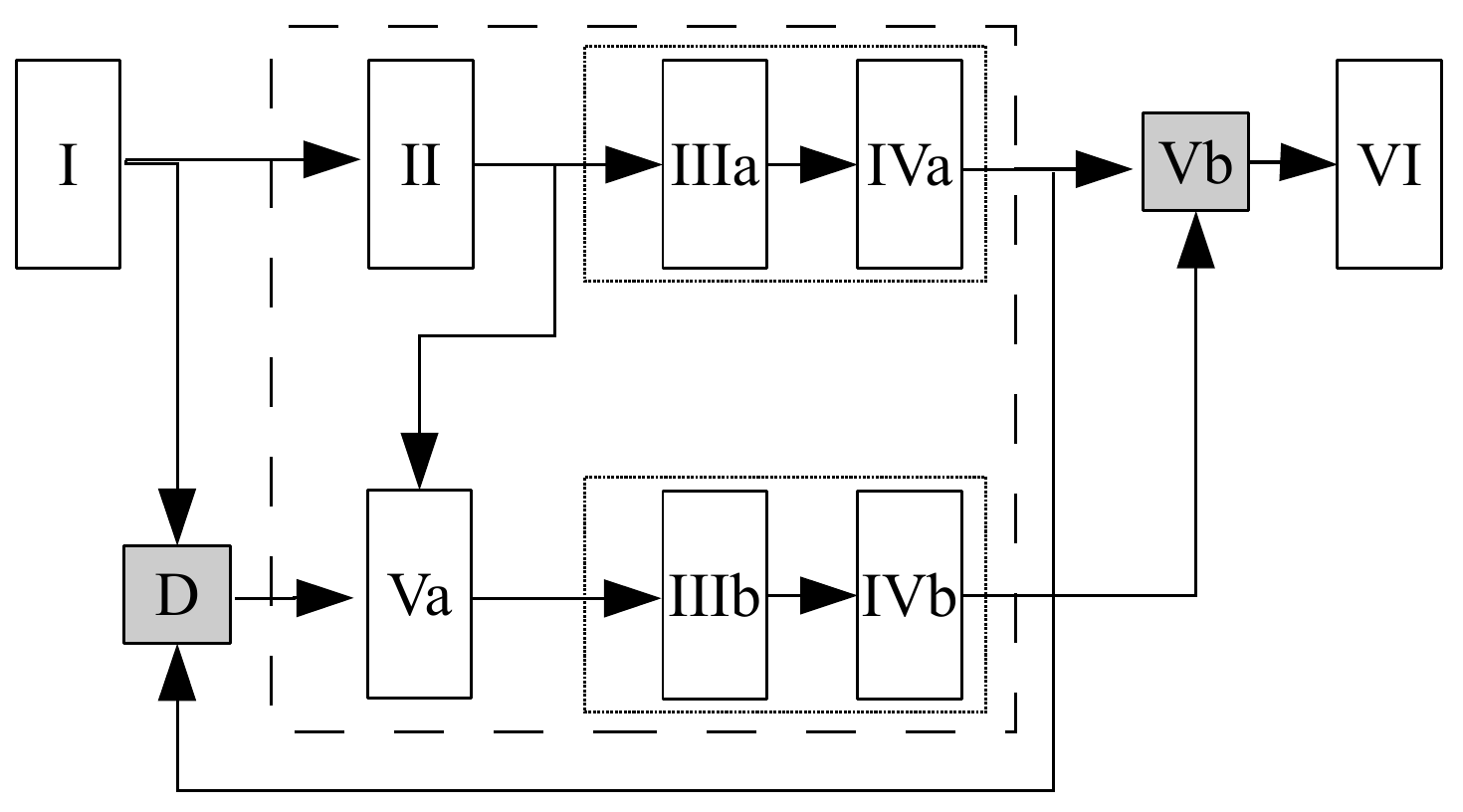}
      \caption{An equivalent recurrent schema}
      \label{fig:recurrent}
    \end{minipage}
\end{figure}

Fig.~\ref{fig:cascade} represents the proposed architecture, which
will be detailed in the following.  It is possible to recognise two
groups of modules, the first comprising \texttt{IIIa} and
\texttt{IVa}, whereas the second \texttt{IIIb} and \texttt{IVb}, each
acting as a FFNN.
The proposed novel topology behaves as a cascade FFNN topology \cite{Schetinin}.
Fig.~\ref{fig:recurrent} depicts a more complex novel topology that
performs the prediction as a Nonlinear AutoRecoursive with exogenous
inputs (NARX) recurrent neural network topology \cite{Zipser}.  Such figure shows
the implemented delay lines to the blocks performing the neural
processing. It should be noted that we have implemented one neuron as
a \emph{purelin} while the remaining neurons in the first hidden layer
process the input signal.  The performed simulations have shown an increased
computational effort, for this recurrent scheme, while the
corresponding results have not significantly improved the accuracy on
the predicted data. Even though this is a novel recurrent cascade
topology this paper fully 
investigates the scheme shown in Fig.~\ref{fig:cascade}.
The following provides the details of the proposed NN cascade.

\paragraph{Input data analysis.}

The input layer (\texttt{I}) does not directly provide the input
vector ($\mathbf{u}$) to the first FFNN hidden layer (\texttt{IIIa}),
being it firstly processed by an intermediate layer (\texttt{II})
that is trained to extrapolate a set of parameters necessary to
perform the $\omega$-validation, i.e.\ the $\sigma$ for the Gaussian
window Fourier analysis. This layer (\texttt{II}) is also provided
with ad adjunct \emph{purelin} neuron acting as a transmission line
for the following layer (\texttt{IIIa}).  

The main purpose of \texttt{II} is to characterise the frequency peaks
windows on the data spectrum in order to associate, after the training
phase, an optimum $\sigma$ value to perform gaussian-window Fourier
analysis on the output data from the second FFNN hidden layer
(\texttt{IVa}).  For this reason, the input to the following layers
are provided as

\begin{equation}
\begin{array}{rll}
\mathbf{x}^{\texttt{Va}} =&\mathbf{y}^{\texttt{II}}(\mathbf{x}^{\texttt{I}}) =& [\tau, \Delta_\tau, \sigma \vert \mathbf{x}^{\texttt{I}} ]\\ \\ 
\mathbf{x}^{\texttt{IIIa}} =& \mathbf{x}^{\texttt{I}}&~\\
\label{eq:io1}
\end{array}
\end{equation}
where $\mathbf{x}^{\texttt{Va}}$ retains the discrete sample number
$\tau$ and both the window size $\Delta_\tau$ and $\sigma$ for the
described Fourier analysis. 

\paragraph{FFNNs hidden layers} The first neuroprocessing module acts
as a fully connected FFNN and consists of two hidden layers, i.e.\
\texttt{IIIa} and \texttt{IVa}.  The first hidden layer
(\texttt{IIIa}) embeds 10 neurons with \emph{tansig} activation
function, whereas the second hidden layer (\texttt{IVa}) consists of 7
neurons with \emph{logsig} activation function.
Similarly, the second neuroprocessing module provides the
functionalities of a fully connected FFNN, however its two hidden
layers, \texttt{IIIb} and \texttt{IVb}, consist of 8 and 5 neurons
with \emph{tansig} activation function, respectively.

\paragraph{FFNN training and validation} The implemented FFNN
neuroprocessing modules are trained by the Levenberg-Marquardt
algorithm with a gradient descent method.  Hence, for the $\tau$-esime
discrete time step, the variation introduced to the weights are given
by
\begin{equation}
\begin{array}{rl}
w^{\mu\nu}_{ij} (\tau)=& w^{\mu\nu}_{ij} (\tau-1) -\eta
e(\tau)\frac{\partial e(\tau)}{\partial w^{\mu\nu}_{ij}(\tau)} \\ \\  
e^\mu(\tau) =& \tilde{\mathbf{y}}^\mu(\tau)-\mathbf{y}^\mu(\tau)
\end{array}
\end{equation}
where  $w^{\mu\nu}_{ij} (\tau)$ represents the value for the
$\tau$-esime step of the connection weight from the $i$-esime neuron
of the $\mu$ layer to the $j$-esime neuron of the $\nu$ layer, $\eta$
is the learning rate parameter, $\tilde{\mathbf{y}}^\mu(\tau)$ and
$\mathbf{y}^\mu(\tau)$ are respectively the training and output signal
from the $\mu$ layer.   
 
\paragraph{$\omega$-validation} The output of the first
neuroprocessing module comes from the second FFNN hidden layer
(\texttt{IVa}) and is  sent, as valid output, to the last layer of the
network and also as input to the validation module (\texttt{Va}). The
validation module consists of a functional unit performing the fast
Fourier transform on a selected window of the input signals. Moreover,
the validation module uses  a dynamically allocated buffer to
implement a size-varying delay line.  

The latter is used to enable real-time online resizing of the Fourier
window to suit the properties of the investigated signal. These
adjustments are performed starting from the parameters contained in
$\mathbf{x}^{\texttt{Va}}$ \eqref{eq:io1}.  Once the gaussian windowed
Fourier transform has been computed, the following values are determined 

\begin{equation}
\begin{array}{rl}
M(\tau,\Delta_\tau,\sigma) =&\max\limits_{[\tau:\tau+\Delta_\tau]}\left\{~\left\vert \hat{F}_\sigma[\tilde{\mathbf{y}^{\texttt{IVa}}}] - \hat{F}_\sigma[\mathbf{y}^{\texttt{IVa}}]\right\vert~\right\}\\
m(\tau,\Delta_\tau,\sigma) =&\min\limits_{[\tau:\tau+\Delta_\tau]}\left\{~\left\vert\hat{F}_\sigma[\tilde{\mathbf{y}^{\texttt{IVa}}}] - \hat{F}_\sigma[\mathbf{y}^{\texttt{IVa}}]\right\vert~\right\}
\end{array}
\label{eq:Mm}
\end{equation}

then the module is trained to admit only certain regions of the
$(M,m)$ pairs plan which validate the output signal of the layer
\texttt{IVa}.

If the output signal results validated, it is then sent as input for the layer \texttt{IIIb} as 

\begin{equation}
\mathbf{x}^{\texttt{IIIb}} = [\mathbf{x}^{\texttt{I}} | \mathbf{y}^{\texttt{IVa}} ]
\end{equation}

The first neuroprocessing module takes $\mathbf{x}^{\texttt{I}}$ as
input and is trained by all the available patterns, while the
second module is trained only by the allowed sequences selected
according to the validation procedure. In the other case, i.e.\ if the
$\omega$-validation is negative, the second module skips the data
during the training process and gives a NaN flagging, being the
relative data for the second variable  unavailable.

\paragraph{Final output} Finally, the implemented topology gives a
global output with a layer consisting of two neurons \emph{purelin}.

\section{Training procedure on OpenMP}

The neural network architecture proposed above has introduced a
sequential validation phase for the results of the first
neuroprocessing module. Validation has to be performed before the first
module results can be sent as input for the second module. Unfortunately,
such sequential operations make the training process expensive
in terms of CPU time.
In order to shorten training time, this section describes a parallel
implementation of the same neural network architecture, using OpenMP,
that manages to obtain asynchronous training and validation.

Generally, when parallelising an application using OpenMP, processes
are forked, joined and synchronised (e.g.\ by means of a barrier).
Such mechanisms, however, introduce a runtime overhead, e.g.\ when
the processes having produced and communicated their results have to
wait until the synchronisation barrier is over.  This is often the
case when the computation times of processes are not perfectly
balanced \cite{OpenMP}.
Therefore, our parallel version aims at reducing such an overhead by
avoiding, as much as possible, the fork-join-barrier constructs, and
by introducing instead processes that produce and consume data.
The main reason for using OpenMP is that, by means of a shared memory,
communication overhead among processes can be avoided, however, on the
other hand, shared memory requires a complex handling of semaphores
and locks before accessing some parts of the memory itself.  We have
handled the synchronisation concern in such a way that overhead is
minimised \cite{Chapman}.

Mainly, the proposed parallel solution is based on the continuous
execution of different processes to care for the different phases of
training for  the above NN cascade.
In our experiments a multi-core processor has been used, however any
kind of shared memory system supporting OpenMP directives can be
employed.

The proposed NN cascade has been trained to predict the values of
$\lambda_{SPP}$ and $L_{SPP}$ starting from an input vector

\begin{equation}
\mathbf{u}(\tau)=\left[ \lambda_0, t \right]
\end{equation}

To evaluate the performance of the NN cascade, two different kinds of
error were considered. We define two \emph{local errors}
$e^{\texttt{a}}$ and $e^{\texttt{b}}$, as well as a \emph{global error}
$e^{*}$ as follows: 

\begin{equation}
\begin{array}{rl}
e^{\texttt{a}}=&\tilde{\mathbf{y}}^{\texttt{IVa}}-\mathbf{y}^{\texttt{IVa}}\\
e^{\texttt{b}}=&\tilde{\mathbf{y}}^{\texttt{IVb}}-\mathbf{y}^{\texttt{IVb}}\\
e^{*}=&\max\left\{ e^{\texttt{a}}, e^{\texttt{b}} \right\} \geq \left | \tilde{\mathbf{y}}^{\texttt{VI}}-\mathbf{y}^{\texttt{VI}} \right |
\end{array}
\label{eq:eee}
\end{equation}
where $\tilde{\mathbf{y}}$ indicates the training value. 

\begin{figure}[!t]
\centering
\begin{minipage}[l]{.48\textwidth}
\centering
\hfill\includegraphics[width=\textwidth]{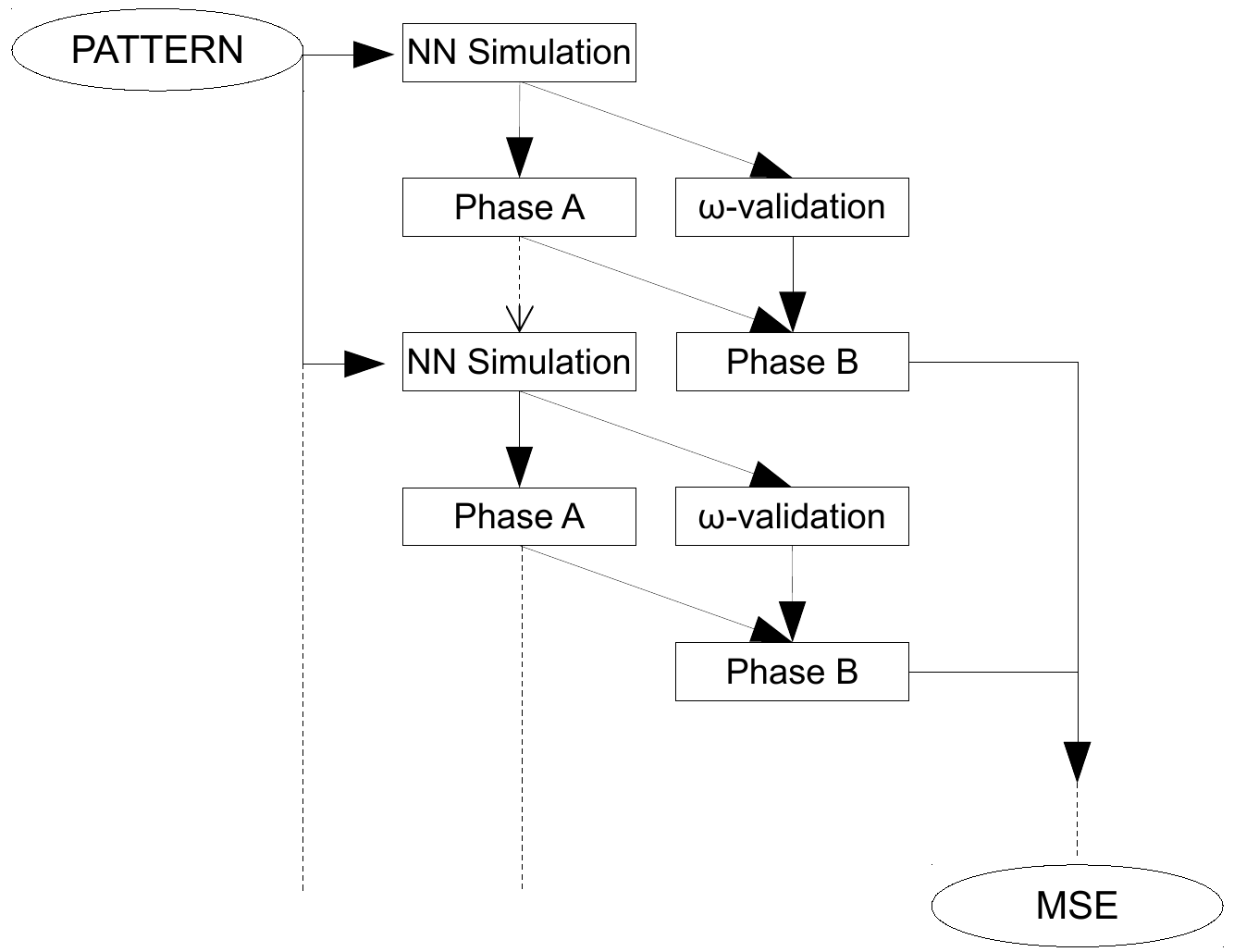}
\caption{The proposed OpenMP training asynchronous stream}
\label{fig:procs}
\end{minipage}
~~
\begin{minipage}[r]{.45\textwidth}
\centering
\hfill\includegraphics[width=\textwidth]{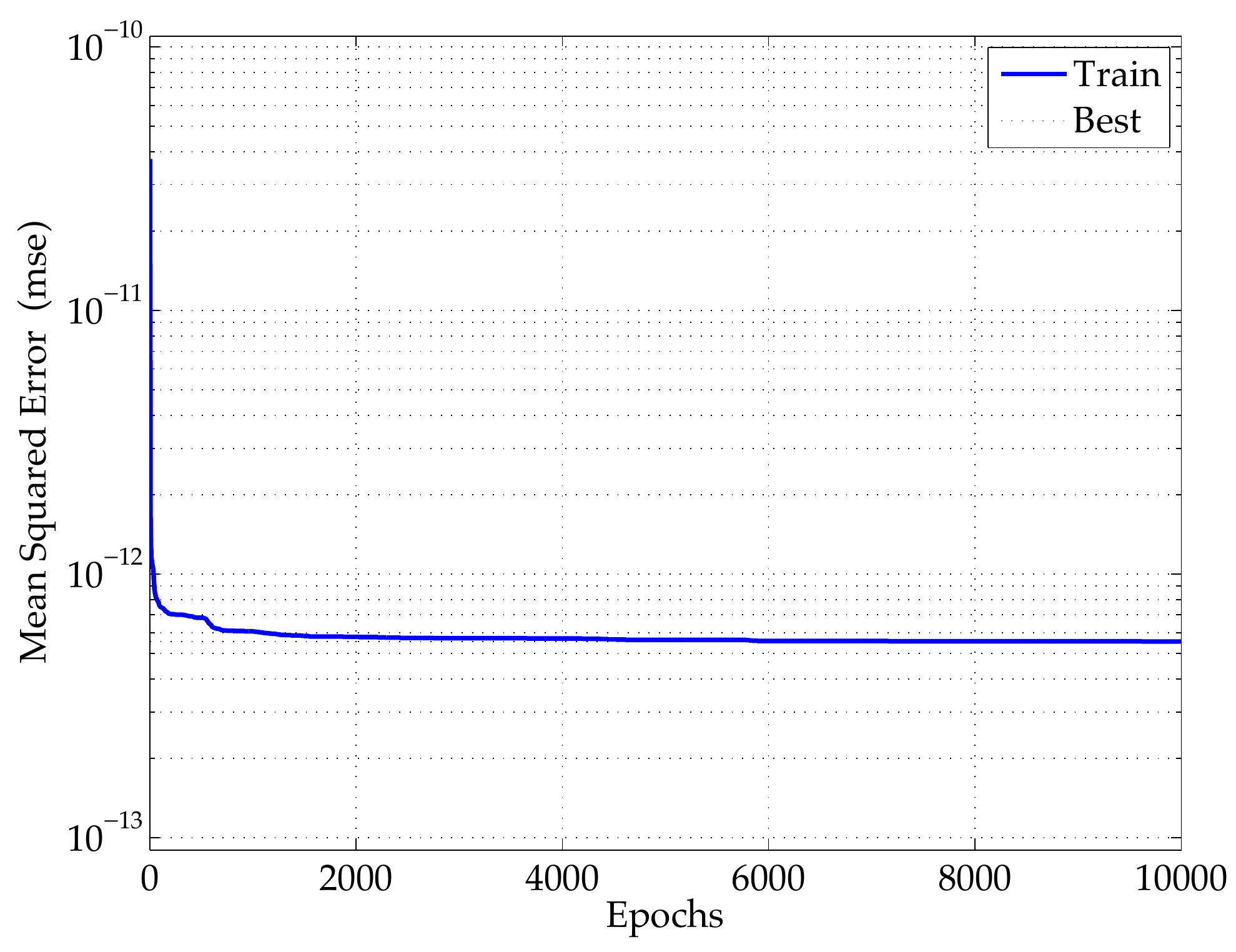}
\caption{Global performance graph of the implemented NN architecture}
\label{fig:perf}
\end{minipage}
\end{figure}

For each training epoch, the outputs from  layers \texttt{IVa},
\texttt{IVb} and \texttt{VI} (see Fig.~\ref{fig:cascade}.)  were used
to compute the errors $e^a$, 
$e^b$ and $e^{*}$ as in \eqref{eq:eee}.
The training 
has been organised in four different activities, executed on an OpenMP
environment (see Fig.~\ref{fig:procs}).

The first activity, named \textsf{NN Simulation}, feeds the whole
neural network cascade with a training pattern, which has been
previously generated.

The second activity, named \textsf{Phase A}, and started once the
first activity has terminated, uses a gradient descent algorithm to
adjust the neural  weights of the intermediate layer \texttt{II} and
the first neuroprocessing (layers \texttt{IIIa} and \texttt{IVa}).

The third activity is 
the \textsf{$\omega$-validation} and is started concurrently with
\textsf{Phase A}, hence after \textsf{NN Simulation} has finished,
since
the results produced by \texttt{IVa} are needed.
The \textsf{$\omega$-validation} activity performs the gaussian windowed fast
Fourier transform of the training set and the predicted signal
resulting as output of \texttt{IVa}, then $M$ and $m$ defined in
\eqref{eq:Mm} are computed.  Eventually, the values of $M$ and $m$ are
used to decide if the pattern data are usable to train the second
neuro-processing module (\texttt{IIIb} and \texttt{IVb}).

Finally, the fourth activity is \texttt{Phase B} performing a further
training that adjusts the output weights of layer \texttt{VI}.
For the proposed schema (see Fig.~\ref{fig:cascade}), module \texttt{Vb}
acts as a controller determining whether it is appropriate to merge
data from \texttt{IVa} and \texttt{IVb} before they can be given as
input to \texttt{VI}.  The merge is enabled when the
\textsf{$\omega$-validation} has given a positive result, otherwise only data
resulting from \texttt{IVa} are used.
Moreover, all the weights in layer \texttt{VI} are adjusted when the result
of  \textsf{$\omega$-validation} is positive, otherwise only the synaptic
weights of the first neuron in \texttt{VI} is adjusted.

The four activities above are started each as a process (see
Fig.~\ref{fig:procs}).  Process \texttt{NN Simulation} feeds data and triggers
the execution of processes \texttt{Phase A} and \texttt{$\omega$-validation}.
The latter two processes give their outputs to process \texttt{Phase B}, and
then \emph{wait} for new data, till the training stops.
Process \texttt{Phase B} starts as soon as input data are available.
At the end of the training epoch the global network performances are
stored for further analysis.  All the measures of performance involved
in the training process are given by the Mean Squared Error (MSE),
though for the global network performances, the formula is adjusted
by using the global error $e^{*}$ of \eqref{eq:eee}.

Fig.~\ref{fig:procs} shows in two vertical tiers some rectangles.
Each rectangle corresponds to a process that can execute in parallel
with another that is on the same row.  In the picture, the time
evolves while going down.  The arrows with continuous lines represent
a flow of data from a producer to a consumer process, whereas the
dotted line the communication of an event.
Ellipses show repositories of data. 
The said interactions among processes are iterated until the training
session stops.

While having devised a parallel solution, our effort has been to
optimise the use of computational resources, hence autonomous
processes needing as less synchronisation as possible have been
implemented as described above.  Our proposed solution manages to
greatly reduce the wall-clock timeframe needed for the training.
 
\section{Results and conclusions}

\begin{figure}[!t]
\centering
\begin{minipage}[c]{.5\textwidth}
\centering
\hfill\includegraphics[width=\textwidth]{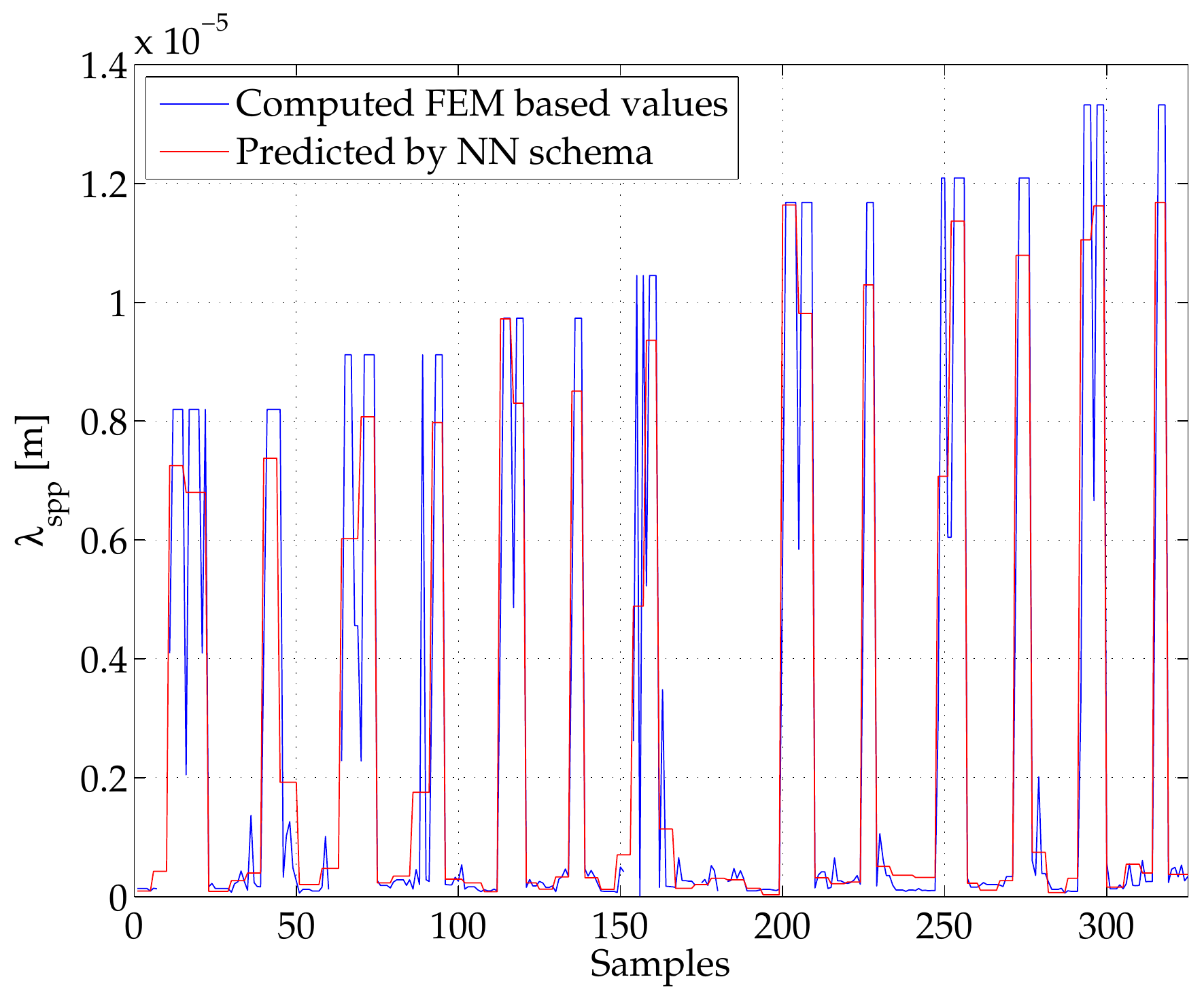}
\end{minipage}
\begin{minipage}[c]{.5\textwidth}
\centering
\hfill\includegraphics[width=\textwidth]{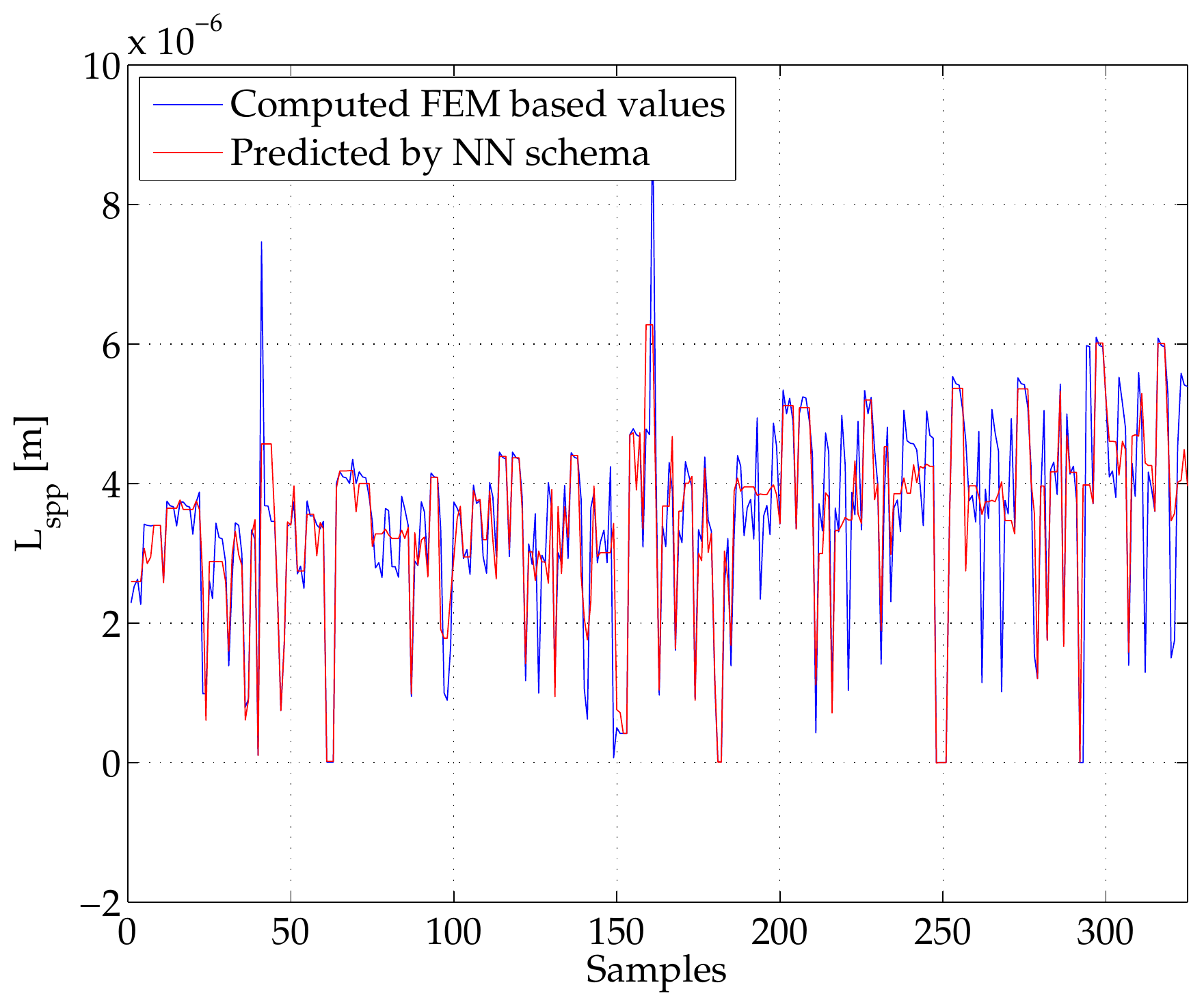}
\end{minipage}
\caption{The computed and predicted $\lambda_{SPP}$ and $L_{spp}$}
\label{fig:out1}
\end{figure}

For training and evaluation we have used the global error $e^*$ to
compute the mean square error (MSE) of the
network. Fig.~\ref{fig:perf} shows the performance of the proposed and
implemented novel cascade NN architecture in terms of such
metrics. Fig.~\ref{fig:out1} reports the values of the computed and
predicted $\lambda_{SPP}$ and $L_{SPP}$. The obtained results confirm
the good predictions obtained by the novel NN schema. 

The proposed NN cascade has been mainly derived from a couple of
common two-layer feed-forward neural networks used to separately
predict $\lambda_{SPP}$ and $L_{SPP}$. The comprehensive structure is
similar to a cascade feed-forward, where the output of the first
neuroprocessing stage has been connected with the inputs for the
second stage to form a new extended input vector.

Simulation results for the NN cascade confirm the effectiveness of the
developed novel architecture whose performance during the training and
evaluation phases show a very low MSE.  Other complex NN architectures
such as pure NARX model or advanced Wavelet Recurrent Neural
Networks~\cite{Napoli} could not be used because of the prediction
instability for the data at hand.

\end{document}